\documentclass[10pt,twocolumn,letterpaper]{article}

\usepackage{cvpr}
\usepackage[dvipsnames]{xcolor}

\definecolor{cvprblue}{rgb}{0.21,0.49,0.74}
\usepackage[pagebackref,breaklinks,colorlinks,citecolor=cvprblue]{hyperref}

\usepackage{booktabs}
\usepackage{graphicx}
\usepackage[accsupp]{axessibility}
\usepackage{multirow}

\def\confName{CVPR}
\def\confYear{2024}

\title{Coarse or Fine? Recognising Action End States without Labels} 

\author{
Davide Moltisanti\thanks{Work done while at the University of Edinburgh}\\
University of Bath\\
{\tt\small dm2460@bath.ac.uk}
\and
Hakan Bilen \hspace{25pt} Laura Sevilla-Lara \hspace{25pt} Frank Keller\\
The University of Edinburgh\\
{\tt\small \{h.bilen, l.sevilla, frank.keller\}@ed.ac.uk}
}

\begin{document}
\maketitle

\begin{abstract}
We focus on the problem of recognising the end state of an action in an image, which is critical for understanding what action is performed and in which manner. 
We study this focusing on the task of predicting the coarseness of a cut, i.e., deciding whether an object was cut ``coarsely'' or ``finely''.
No dataset with these annotated end states is available, so we propose an augmentation method to synthesise training data. We apply this method to cutting actions extracted from an existing action recognition dataset. Our method is object agnostic, i.e., it presupposes the location of the object but not its identity. 
Starting from less than a hundred images of a whole object, we can generate several thousands images simulating visually diverse cuts of different coarseness.
We use our synthetic data to train a model based on UNet and test it on real images showing coarsely/finely cut objects. 
Results demonstrate that the model successfully recognises the end state of the cutting action despite the domain gap between training and testing, and that the model generalises well to unseen objects.
\end{abstract}

\section{Introduction}
\label{sec:intro}

Action recognition is central to understanding the visual world. 
When we observe people performing a task (e.g.,~cooking dinner), we are able to decompose the task in terms of discrete actions (e.g.,~boiling water, cooking pasta). AI systems that understand videos 
face the same problem and in response to this, a large literature on action recognition was sprung up~\cite{kong2022human}. 
A system recognising actions needs to identify the people and objects involved in the action. Crucially, most actions 
are characterised by the \textbf{state change} of an object. An example is the \textit{cutting} action: independent of which object this action is applied to, it will result in the object being in smaller parts than before. Moreover, the \textbf{manner} expressed by an adverb \cite{Doughty_2020_CVPR,Doughty_2022_CVPR,Moltisanti_2023_CVPR} 
in which an action is performed is crucial to understanding it. 
For instance, 
to cut garlic \textit{finely} we would perform a different operation (e.g., we would mince it with a knife) compared to what we would do to cut it \textit{coarsely} (e.g., we might just split it with our hands).
Recognising object end states as the result of the way an action is performed is thus important to enable systems to understand more about the action itself. This is a hard endeavour because even for a single action 
there are a large number of objects that the action can be applied to. Not only do these objects vary visually, 
the end state will also look differently depending on the type of object and the manner of the action. For example, a finely cut carrot would typically be sliced in small strips, whereas a finely cut clove of garlic would be minced. 

\begin{figure}[t]
    \centering
    \includegraphics[width=\linewidth]{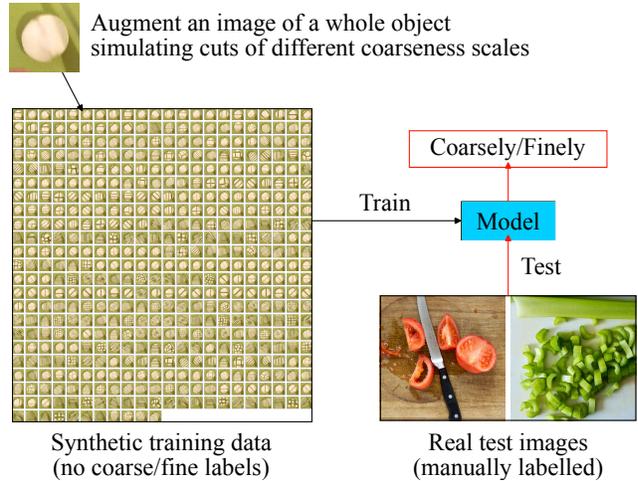}
    \caption{Summary of our work. We aim to recognise the end state of an action, e.g., whether an object is cut coarsely or finely. We assume no labels and propose an object-agnostic image augmentation method to synthesise training data. Our model successfully learns from this synthetic data, as we show by testing on real images and videos, including for unseen objects.}
    \label{fig:teaser}
\end{figure}

In this paper, we present a case study to recognise the end states of one action, cutting, characterised by the main manner in which it can be performed: \textit{coarsely} and \textit{finely}. 
To focus on the end state recognition problem we do not work on videos and assume only an image depicting the end of the action is given. This allows us to isolate the task of discerning a coarse cut from a fine cut, without having to worry about video-related issues such as motion, finding the object in a video, etc. 
We propose a method to generate training data to address the limited availability of datasets labelling coarsely/finely cut objects in images.
Using \textbf{data augmentation,} we generate a large, high-quality dataset of synthetic images (VOST-AUG) on which we then train our model for end state recognition. This is illustrated in Figure~\ref{fig:teaser}.
Our augmentation method starts with an image of a whole object and a mask segmenting the object. The approach is object agnostic in that it does not need to know \textit{what} object is in the picture, only \textit{where} it is. We devise several ways to control the simulated coarseness of a cut, which enables us to generate numerous diverse images from a single source. Starting from only 184 images, we generate 90,809 images simulating objects being cut in realistic ways with different coarseness. 
We also propose a model based on a UNet~\cite{ronneberger2015u} Encoder-Decoder architecture to take full advantage of our data. Our model is able to achieve 0.856 Mean Average Precision (MAP) on unseen objects, a 4\% improvement over the closest baseline.

Our training data is synthetically generated, 
so it is crucial to test whether the model we propose is able to generalise to real images and videos. Our experiments demonstrate that this is the case, both for an image test set that we collected and annotated (COFICUT) and for the existing video dataset Adverbs in Recipes ~\cite{Moltisanti_2023_CVPR}. In both cases, our model outperforms an existing adverb recognition model \cite{Moltisanti_2023_CVPR}. Furthermore, we show that our approach beats a supervised model trained on a portion of COFICUT.

To summarise: (i)~We focus on the problem of recognising the end state of an action, which is critical for both action and manner (adverb) recognition. (ii)~We propose an object-agnostic augmentation method to synthesise training data for this task. (iii)~We present a model based on UNet and train it on our synthetic data. (iv)~To evaluate the effectiveness of the synthetic images, we collect a small test set of real images showing coarsely/finely cut objects. Both on this test set and on an existing video dataset, our model achieves good performance, even for unseen objects.

\vspace{-7pt}
\section{Related Work}
\label{sec:related_work} 

\paragraph{Adverb Recognition in Videos}

The closest line of research to action end state recognition is understanding adverbs in videos~\cite{Doughty_2020_CVPR,Doughty_2022_CVPR,Moltisanti_2023_CVPR,hummel2023video}, where models learn to recognise the manner in which actions are performed in a video. In some cases this includes end states, as in ``cut coarsely/finely''. 
The approaches of \cite{Doughty_2020_CVPR,Moltisanti_2023_CVPR,hummel2023video} are fully-supervised, while~\cite{Doughty_2022_CVPR} proposes a method that assigns pseudo-labels to the training videos based on the model predictions. However, this still assumes adverb labels are available, since pseudo-labels are assigned from the set of classes in a given dataset. In existing datasets for adverb recognition~\cite{Doughty_2020_CVPR, Doughty_2022_CVPR,Moltisanti_2023_CVPR} videos are loosely trimmed and often noisy, without a ground truth localising which frames show the object. This means that learning action end states from such datasets would be difficult.
In contrast, we generate training images via augmentation without action, adverb, or object labels. Our model learns to recognise the end state of an action from augmented images in a granular way, i.e., without splitting images into adverb categories. Nevertheless, we show that our model outperforms the adverb recognition model presented in~\cite{Moltisanti_2023_CVPR}, including on videos from the Adverbs in Recipes dataset~\cite{Moltisanti_2023_CVPR}.

\vspace{-10pt}
\paragraph{Object Attributes in Images} Our task also overlaps with the problem of predicting attributes in images~\cite{wang2010discriminative,wang2013unified,chen2014inferring,isola2015discovering,misra2017red,nagarajan2018attributes,nan2019recognizing,li2020symmetry,mancini2021open,naeem2021learning,saini2022disentangling,wang2023learning}, with an important distinction: in object attribute discovery images are typically grouped in a \textit{single} category (e.g.,~tomato), and the goal is to organise the input group of images into distinct states or attributes (e.g.,~ripe, raw, peeled, etc). We instead start from an unstructured group of objects and aim to recognise a change in a visual attribute: the coarseness resulting from a cut.
In other cases object attribute discovery is addressed from a zero-shot compositional learning perspective, which is a distinct problem compared to action end state recognition. Nevertheless, for completeness we also adapt an attribute discovery model~\cite{wang2023learning} when comparing to state-of-the-art work.
We note that the popular 
MIT-States~\cite{isola2015discovering} annotates adjectives 
including ``cut, sliced, peeled, chopped'' and especially ``thin/thick''. 
However in this dataset ``thin/thick'' do not necessarily correspond to ``coarsely/finely cut'', i.e., there are objects such as ``sauce, cloud, wall, book, etc'' annotated with ``thin/thick''. For this reason this dataset is not a suitable resource for our problem. 

\vspace{-10pt}
\paragraph{Image Augmentation} The success of deep learning on image tasks is in good part due to image augmentation techniques such as cropping, rotation, colour and perspective modifications, etc~\cite{shorten2019survey,mumuni2022data}. Indeed, thanks to these techniques we can expand the visual and semantic diversity of the training data to prevent models from overfitting and enhance their generalisability. In this work we propose a method to augment images, however our method is tailored to synthesise training data from the scratch rather than augmenting an existing training dataset.

\vspace{-5pt}
\section{Probing Existing Methods}
\vspace{-2pt}
\label{sec:probing}

\begin{figure*}[t]
    \centering
    \includegraphics[width=\linewidth]{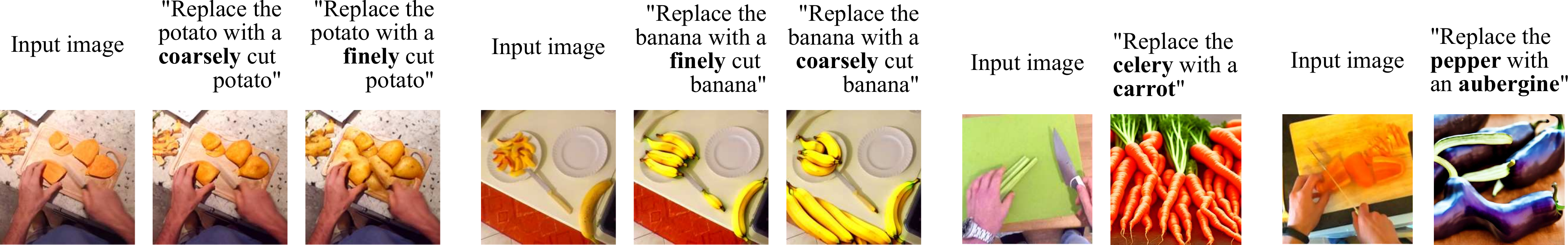}
    \caption{Trying to generate images of coarsely/finely cut objects with InstructPix2Pix~\cite{brooks2023instructpix2pix}. Text indicates the prompts used.}
    \label{fig:probing}
\end{figure*}

In this section we will try to establish how good current retrieval systems are at telling if an object was cut coarsely or finely. We search for food images on Microsoft Bing using the query ``\{coarsely, finely\} cut $o$'', where $o$ is one of 27 objects such as ``carrot, garlic, tomato'' (see Section~\ref{sec:a-coficut-details} for the full list).
We take the top 100 retrieved images, drop duplicates and inspect each image to establish how many images were incorrectly retrieved, i.e., showing a coarse cut when searching for a fine cut, and vice-versa. Amongst 1,869 images, we found that 42.5\% were incorrectly retrieved. This high percentage suggests that retrieval models struggle to distinguish images in these two categories. 
We do not have internal access to the retrieval system employed by Microsoft Bing, i.e., we do not know for sure whether the search engine uses vision-based text-image retrieval models. If this is the case, then we can ascribe the relatively poor coarse/fine retrieval performance to the fact that text-image retrieval models are typically optimised to distinguish objects into broad classes rather than fine-grained categories. 
We believe the main reasons for this are the lack of extensive fine-grained labels and the fact that models need to learn beyond the visual appearance of an object, i.e., they need to be able to generalise to recognise coarseness across visually distinct objects.  

Generative models are often used in low data regimes to synthesise new images with the desired label. We thus experimented with a generative model to synthesise images of coarsely/finely cut objects. We tested InstructPix2Pix~\cite{brooks2023instructpix2pix}, which uses GPT-3~\cite{brown2020language} and Stable Diffusion~\cite{rombach2022high} to edit images based on a text prompt. For testing, we used a few images from EPIC Kitchens~\cite{damen2022rescaling}, asking the model to replace a visible object with a finely/coarsely cut version of the same object. 
Figure~\ref{fig:probing} shows some examples from this experiment (first two columns). The model generates mostly plausible images, however it just replaces the prompted object with a newly generated version of the same object, ignoring the adverb in the prompt. 
InstructPix2Pix was trained with hand-made transformation prompts, which means our test instructions are too distinct from the prompts the model was trained on. We further probe this with simpler prompts, asking the model to only replace an object with another one. We still see that the model fails (last two columns) despite the easier prompts. This confirms that the model has not been trained to cover our domain of interest sufficiently. 
We show more examples in Section~\ref{sec:a-more-examples-p2p}.

From this Section we conclude that current retrieval methods struggle to differentiate a coarsely cut object from a finely cut one, and that current generative models cannot reliably synthesise images of coarsely/finely cut objects. 
We therefore propose an image augmentation method that makes it possible to generate synthetic images of coarsely and finely cut objects. We show that this synthetic data can be used successfully to train a classifier that works on real (not synthesised) images of coarsely/finely cut objects.

\section{Dataset Creation}
\label{sec:augmenting_images}

\begin{figure*}[t]
    \centering
    \includegraphics[width=\linewidth]{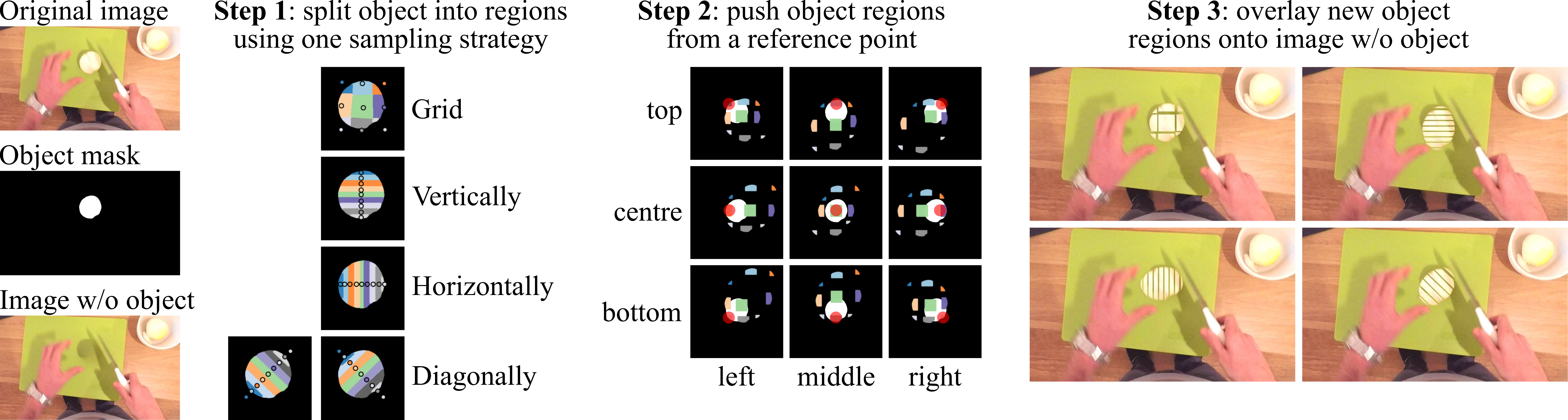}
    \caption{Our augmentation method to transform whole objects into cut objects. Given an image and a mask segmenting the object, we first remove the object and inpaint the image to fill the resulting hole (image w/o object, bottom left). We then split the object into regions (\textbf{Step 1}). For this we sample $n$ seeding points (nine in this example, indicated by circles) and group object pixels into regions based on their distance to each point, as in a Voronoi diagram. We devise four sampling strategies which affect the topology of the regions and simulate different cut types. 
    We then ``break'' regions given a reference point (\textbf{Step 2}), shown as a red dot, i.e., we push each region away from the reference point along the line connecting the region and the point. 
    Lastly (\textbf{Step 3}), we overlay the new regions onto the image w/o object to obtain the final augmented image. We show four examples with reference point (centre, middle) and each of the four sampling strategies.}
    \label{fig:aug_method}
\end{figure*}

\begin{figure*}[t]
    \centering
    \includegraphics[width=\linewidth]{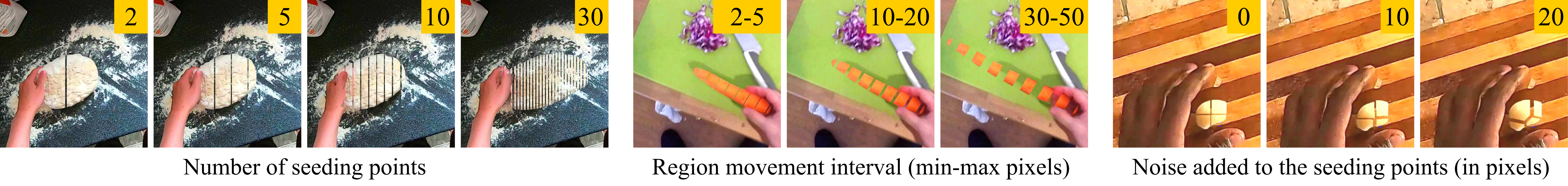}
    \caption{Illustrating how the parameters of our augmentation affect the output image. The number of seeding points controls the coarseness of the simulated cut, with fewer/more points corresponding to a coarser/finer cut (left). To obtain more diversified and realistic images we push regions by a random number of pixels sampled within an interval (centre) and add noise to the seeding points (right).     
    }
    \label{fig:aug_examples}
\end{figure*}

\subsection{Augmenting to Simulate Cuts}

Let us assume we are given an image depicting an object in its whole state and a mask segmenting the object. Our goal is to generate several images depicting the object as if it was cut at different coarseness levels, i.e., from coarsely to finely. We augment the image to achieve this. Specifically, we first remove the object from the image, which we then inpaint to fill the hole left by the removed object using~\cite{dong2022incremental}.
Next we ``break'' the object to simulate the result of a cutting action, and overlay the split parts of the object onto the inpainted image to obtain a picture where the object is cut.

Figure~\ref{fig:aug_method} illustrates our augmentation method in detail. To break the object, we start sampling $n$ points from the mask. 
The sampled points act as seeding points to segment object regions, which are obtained by grouping pixels that are closest to one of the $n$ seed points. This is how Voronoi diagrams are built, with the important difference that points are not random but sampled in a way that simulates different human cuts.
Specifically, we devise four sampling strategies: \textit{grid:} we sample uniformly both horizontally and vertically, which simulates an object being cut in squares or cubes; \textit{horizontally/vertically:} points are sampled only horizontally or vertically, which simulates objects being cut in vertical or horizontal strips; \textit{diagonally,} where points are sampled along the main or secondary diagonal of the mask, which also simulates objects cut in strips but with an angle (see ``Step 1'' in Figure~\ref{fig:aug_method}). Points are evenly spaced initially, however we add random noise to each seeding point to get a more natural looking cut. 
We next move object regions by a few pixels to ``break'' the object (``Step 2'' in Figure~\ref{fig:aug_method}), selecting a reference point and pushing each region along the line connecting the region to the reference point. To generate more natural and diverse images, each region is shifted by a number of pixels randomly sampled within an interval.
Finally, the moved object regions are overlaid onto the inpainted image without object (``Step 3'' in Figure~\ref{fig:aug_method}). 

Figure~\ref{fig:aug_examples} shows a few synthetic images illustrating how the parameters of our augmentation method affect the resulting image. The most important parameter is the number of seeding points, which controls the coarseness of the cut (fewer/more points correspond to a coarser/finer cut). The random region movement and the seeding point noise ensure that images are more diverse and natural-looking. 
Interestingly, these two parameters also affect the perceived \textit{roughness} of a cut, i.e., a greater seeding point noise and a greater region movement will make the cut look rougher or more haphazard. Semantically, the concepts of roughness and coarseness overlap, so it would be difficult and somewhat arbitrary to label the augmented images as coarse or fine based on one or multiple augmentation parameters. For this reason we do not label images into categories. We later show that the difference between an original and augmented image provides a proxy measure to gauge coarseness.

\subsection{The VOST-AUG Dataset}

\begin{table}[t]
\centering
\resizebox{\columnwidth}{!}{%
\begin{tabular}{@{}lc|lc@{}}
\toprule
\textbf{Original images}     & 184    & \textbf{Objects}                    & 41     \\
\textbf{Augmented images}    & 90,809 & \textbf{Objects seen in training}   & 30     \\
\textbf{Avg. aug. per image} & 493    & \textbf{Objects unseen in training} & 11     \\
\textbf{Training orig. images}     & 96 & \textbf{Testing orig. images}             & 84 \\
\textbf{Training aug. images}     & 47,395 & \textbf{Testing aug. images}             & 43,414 \\\bottomrule
\end{tabular}%
}
\caption{Summary of the VOST-AUG dataset. Starting from only 184 images we generate 90,809 augmentations showing objects cut at different coarseness scales.
\vspace{-10pt}
}
\label{tab:vost_aug}
\end{table}

\begin{figure*}[t]
    \centering
    \includegraphics[width=\linewidth]{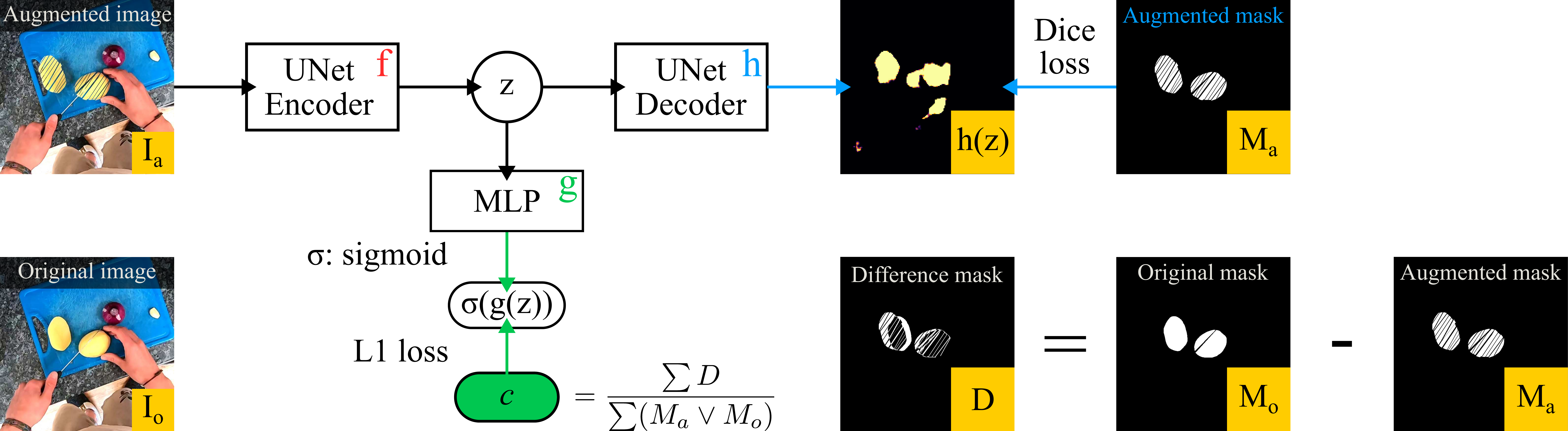}
    \caption{Our model to predict the coarseness of a cut. The model adopts a UNet architecture, where the Encoder bottleneck features $z$ are optimised in two ways. We use an MLP to predict coarseness given $z$ with the L1 loss, using $c$ as target. To learn a stronger $z$, the UNet decoder adds an auxiliary segmentation task, where we use the augmented object mask as target. The decoder is used only during training. For inference we employ only the Encoder and the MLP output to predict the coarseness of a test image.}
    \label{fig:architecture}
\end{figure*}

We now provide details about how we use augmentation to generate a dataset for this work. We are interested in exploring the potential of a small-scale but high-quality set of annotated images, i.e., we would like to see whether it is possible to train a model to recognise a coarse or fine cut starting from a small set of good images. With this premise, the Video Object Segmentation under Transformations (VOST) dataset~\cite{tokmakov2023breaking} is a good resource, as it focuses on actions that significantly transform an object and offers high-quality manual object masks for a few videos from EPIC Kitchens~\cite{damen2022rescaling} and Ego4D~\cite{grauman2022ego4d}. Furthermore, objects are typically well visible in these datasets thanks to the egocentric viewpoint.
VOST annotates 702 videos comprising different actions such as ``cut, squeeze, paint'', etc. We select only videos labelled with the verb ``cut'', obtaining 184 videos 
showing 41 different objects. 
Video segments are well trimmed in EPIC Kitchens and Ego4D, thus we assume that the first frame in each segment contains the object in its whole state and select the first frame of each segment to build our set of images to augment.\footnote{Some segments are part of a repeated action where objects appear partially cut in the first frame. This was not an issue for our augmentations.}

As detailed before, there are a few parameters involved in our augmentation method. For this work we augment images taking all combinations of the following parameter values: number of seeding points: 2, 3, 5, 10, 20, 30, 40, 50; seeding points sampling: `diagonal (main)', `diagonal (secondary)', `grid', `horizontal', `vertical'; region movement intervals: $[2, 5], [5, 10], [10, 20]$; seeding points noise: 0, 5, 10, 20, 50. For each combination we sample a random reference point among the nine illustrated in Figure~\ref{fig:aug_method}. With these combinations we generated in total 90,809 augmented images, starting from only 184 original images. On average there are 493 augmented images per original image (some augmentations are rejected if object regions are pushed outside the image bounds, which can happen if the object is near an edge). We split the augmented images in a 70/30 ratio for training/testing, where all augmented images from a given source are either in the train or the test split. We call the set of augmented images the \textbf{VOST-AUG} dataset. Table~\ref{tab:vost_aug} provides a summary of our dataset. We show more examples in Section~\ref{sec:a-vost-more-details}.

\section{Model}
\label{sec:model}

As discussed in Section~\ref{sec:augmenting_images}, the coarseness of a cut is mainly controlled by the number of parts a whole object is cut into, however other factors such as the distance between parts and the regularity of their shape also influence the perceived coarseness. 
We thus design the model based on the difference between an augmented image and its original source: visually, an augmented image will change less/more if the cut is coarser/finer, as we show in Figure~\ref{fig:aug_examples}. 

To quantify this, 
let $M_a$ and $M_o$ be the 2D binary masks segmenting the object in an augmented image $I_a$ and its original source $I_o$. Let $D = |M_a - M_o|$ be the binary matrix obtained taking the absolute value of the pixel-wise difference between the two masks. 
We can measure how much an augmented image and its original source differ by comparing their masks. Formally, we have:
\begin{equation}
c(M_a, M_o) = \frac{D}{\sum (M_a \lor M_o)} = \frac{\sum |M_a - M_o|}{\sum (M_a \lor M_o)}
\label{eq:change_ratio}
\end{equation}
where the denominator normalises the difference between 0 and 1 and ensures that $c$ is independent of the size of the object. Values of $c$ closer to 0/1 indicate a small/large difference between the augmented and the original image, which in turn correspond to a coarser/finer cut.

With the above definition, 
we can now introduce our model to learn $c$ from $I_a$ to discern the coarseness of a cut. In principle, a model trained with a regression objective such as the L1 loss could be sufficient for this task. However, as we will show in Section~\ref{sec:experiments}, it is hard for a model to solve this task without extra guidance due to the subtle differences between the numerous images augmented from a single source. We thus propose an Encoder-Decoder model based on UNet~\cite{ronneberger2015u}. UNet was designed for medical imaging segmentation, where small-scale details are crucial, thus it is particularly suited for our problem. Our model is depicted in Figure~\ref{fig:architecture}. The Encoder $f$ receives in input an augmented image $I_a$ and outputs the bottleneck features $f(I_a) = z$. We optimise $z$ in two ways: firstly, we feed $z$ to an MLP $g$ which outputs a scalar, and use the L1 loss to learn $c$: $\mathcal{L}_{L1} = |\sigma(g(z)) - c|$, where $\sigma$ is the sigmoid function\footnote{$z$ has shape $(2048, u, v)$, we average along dimensions $u$ and $v$ before feeding it to the MLP.}.
The decoder learns a segmentation mask from the bottleneck features $z$ via skip connections with the Encoder. In our case, the output $h(z)$ is a one-channel image with the same shape as the input. We optimise $h(z)$ with the Dice loss~\cite{milletari2016v}, using the augmented object mask as target: 
\begin{equation}
    \mathcal{L}_{Dice} = 1 - \frac{2\sum\big(M_a \odot h(z)\big) + \epsilon}{\sum\big(M_a + h(z)\big) + \epsilon}
\end{equation}
where $\epsilon$ is a small constant for numerical stability. The combined loss to train our model is the sum of the two losses with equal weight, i.e. $\mathcal{L} = \mathcal{L}_{L1} + \mathcal{L}_{Dice}$. The purpose of the Decoder is to provide an auxiliary segmentation task that strengthens the Encoder representation. As is typical with Encoder-Decoder architectures, the Encoder must generate a high-quality representation for the Decoder to effectively solve the dense segmentation task. In other words, the Decoder and the segmentation task provide the ``extra guidance'' to focus on nuanced differences and learn a better representation for our main task, the coarseness cut estimation. 
We also use the original images and masks for training. In this case $c=0$, whereas the target for $\mathcal{L}_{Dice}$ is $M_o$ instead of $M_a$.
We use the Decoder only during training. For inference we only employ the Encoder $f$ and the MLP $g$ and use the output $\sigma(g(f(x)))$ to predict the coarseness of the cut in the image $x$, where values closer to 0/1 indicate a coarse/fine cut as per $c$'s definition.
Our model is able to learn the differences that define the coarseness of a cut by seeing only a single image at the time rather than both the original and the augmented image. This is advantageous as we do not need a reference image for testing. Importantly, our model is object agnostic, so it does not require object labels and does not need an object mask during inference, which makes the model more useful in a real-world setting. We will evaluate the model on real-world (i.e., non-augmented) data in Section~\ref{sec:coficut}.

\section{Experiments}
\label{sec:experiments}

\paragraph{Implementation Details} We employ ResNet50~\cite{he2016deep} pre-trained on ImageNet~\cite{deng2009imagenet} as our backbone for all experiments and baselines. Models are trained with the ADAM optimiser~\cite{kingma2015adam} for 300 epochs with learning rate $1e-4$, weight decay $5e-5$, batch size 64, dropout 0.1 and no batch normalisation. The MLP in our model 
has one hidden layer. Input images are resized to $224\times224$.
All experiments are conducted on a single 12GB NVIDIA GeForce RTX 3060.

\paragraph{Evaluation Metric} We report Mean Average Precision (MAP) with macro average (the two classes have equal weight). 
We report MAP globally as well as for seen/unseen objects (except on AIR, where we do not have object labels).
For evaluation on VOST-AUG, we group images augmented from the same source and assign them coarse/fine labels based on the median value of $c$. To clarify, let $\mathcal{I}^A = (I_a^i, i=1 \dots N)$ be the sequence of $N$ images augmented from the same source $I_o$, and let $\mathcal{C} = (c(M_a^i, M_o), i=1 \dots N)$ be the sequence containing the $c$ values obtained from the corresponding augmented masks (see Equation~\ref{eq:change_ratio}). We label each augmented image as follows:
\begin{equation}
    y(I_a^i) = \begin{cases}
        0 \quad (coarse) \ &\text{\small if} \quad c(M_a^i, M_o) \le \Tilde{\mathcal{C}} \\
        1 \quad (fine) \ &\text{\small if} \quad c(M_a^i, M_o) > \Tilde{\mathcal{C}}
    \end{cases}
    \label{eq:labelling}
\end{equation}
where $\Tilde{\mathcal{C}}$ denotes the median of $\mathcal{C}$.

\paragraph{Baselines} To the best of our knowledge no prior work has focused on our problem with this setting. The closest line of work is adverb recognition in video~\cite{Doughty_2022_CVPR,Moltisanti_2023_CVPR,hummel2023video}. We compare against~\cite{Moltisanti_2023_CVPR}, who propose two methods to recognise adverbs termed ``CLS'' and ``REG''
. We adapt this model as follows, using the same backbone we use for our model. For CLS, we label training images as coarse/fine as we do for testing on VOST-AUG (see Equation~\ref{eq:labelling}). CLS is then a standard classification baseline where the model is trained with binary cross entropy (BCE), i.e., we optimise $\sigma(g(z))$ with the BCE loss (see Figure~\ref{fig:architecture}). 
We also train the CLS model by splitting images based on the number of seeding points instead of the $c$ value, i.e., in Equation ~\ref{eq:labelling} we replace $\mathcal{C}$ with $\mathcal{S} = (s_i, i=1 \dots N)$ and $c(M_a^i, M_o)$ with $s_i$, where $s_i$ is the number of seeding points used to generate the $i$-th augmented image. Test images in VOST-AUG are still split as in Equation~\ref{eq:labelling} to compare all models equally.

For REG in~\cite{Moltisanti_2023_CVPR} verb-adverb video-text embeddings are used to build a regression target. This is sensible when verbs vary, i.e., when there are samples annotated with different verbs for a given adverb. This is not the case in our setting as we only have one verb (cut). 
We thus adapt REG by using the $c$ values as regression target. This is essentially the same as training our model without the Decoder and the segmentation task, so REG serves also as an ablation study for our full model.
All models are trained on VOST-AUG. As we only have two classes, we also provide a random baseline to provide a lower bound. In this case the mean average precision equals the support size of the positive class (testing images with the ``fine'' label).

We also adapt CANet-CZSL~\cite{wang2023learning}, a model for compositional zero-shot attribute learning. We fine-tune the model pre-trained on MIT-States~\cite{isola2015discovering}, training the model to recognise two attributes: ``coarse'' and ``fine''.

\subsection{Datasets}
\label{sec:coficut}

\begin{figure*}[t]
    \centering
    \includegraphics[width=\linewidth]{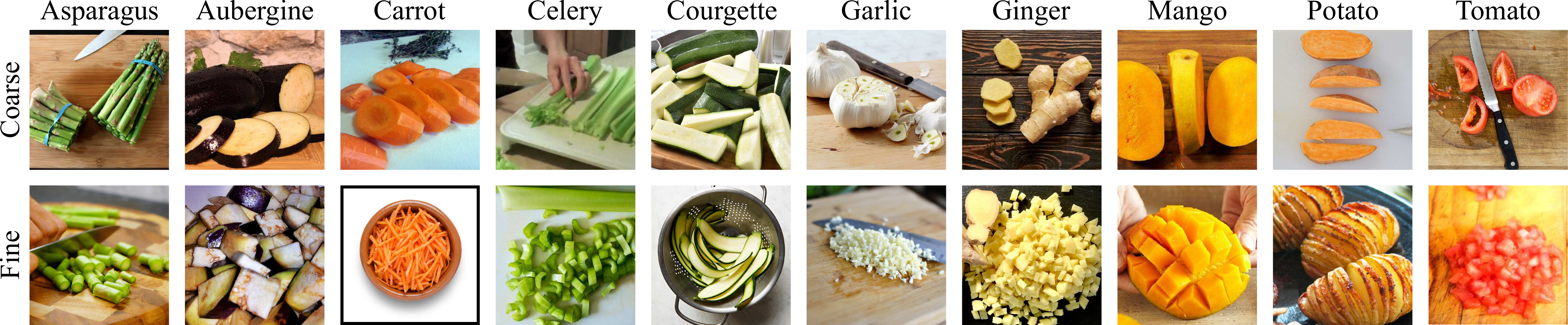}
    \caption{Samples from COFICUT, the dataset of coarsely/finely cut food images we collect for evaluation.}
    \label{fig:coficut_examples}
\end{figure*}

\begin{table}[t]
\centering
\resizebox{\columnwidth}{!}{%
\begin{tabular}{@{}lc|lc@{}}
\toprule
\textbf{Total images}      & 1,869 & \textbf{Objects}                    & 27 \\
\textbf{Finely cut images}   & 1,211 & \textbf{Objects seen in training}   & 19 \\
\textbf{Coarsely cut images} & 658   & \textbf{Objects unseen in training} & 8  \\ \bottomrule
\end{tabular}%
}
\caption{Summary of the COFICUT evaluation dataset.}
\label{tab:coficut}
\end{table}

\paragraph{COFICUT} We collect a set of food images from Microsoft Bing. We start querying ``\{coarsely, finely\} cut $o$'', where $o$ is an object from the list of objects in VOST-AUG, labelling each image with either ``coarse'' or ``fine'' based on the query. We take the top 100 retrieved images. We drop duplicates and manually review all images discarding irrelevant results, adjusting their labels to ensure that each image is correctly annotated (as shown in Section~\ref{sec:probing}, the retrieved images were often relevant to the opposite adverb). We remove objects altogether when there was no visible difference between the coarse and fine images. After reviewing, we retain 1,869 images (1,211 labelled as ``finely'' and 658 labelled as ``coarsely'') showing 27 different objects (of which 8 are not seen in training). We name this dataset \textbf{COFICUT} (COarse-FIne CUT Food Images), which is summarised in Table~\ref{tab:coficut}. This dataset is used only for evaluation. Despite its small scale, images have different viewpoints and style than those seen in training, i.e., in training images are all from a first-person point of view (PoV), whereas in testing they are mostly from a third-person PoV. Training images are daily-life captures, whereas COFICUT images are a mix of product pictures, still frames from vlogs or recipe pictures, with very different lighting and style, as illustrated in Figure~\ref{fig:coficut_examples}. Above all, training images are synthetic augmentations, whereas test images are real examples of cut objects. For these reasons, we believe COFICUT is a challenging benchmark.
COFICUT, VOST-AUG and our code are available at \url{github.com/dmoltisanti/coficut-cvprw24}.

\paragraph{Other Datasets} We also evaluate models on the test split of the VOST-AUG dataset and the video dataset Adverbs in Recipes (AIR)~\cite{Moltisanti_2023_CVPR}. 
As we do not have real binary labels for VOST-AUG, evaluation on VOST-AUG should be seen more as a sanity check rather than a benchmark for comparison. 
AIR annotates 10 adverbs in instructional videos. We select videos labelled with either ``coarsely'' or ``finely'' and one of the following verbs: ``chop, cut, mince, grind, grate'', for a total of 992 videos. Like in COFICUT, the PoV in AIR is different from that in VOST-AUG. Furthermore, the nature of the videos (instructional) introduces additional diversity, e.g., people explaining their actions are often visible, and videos contain jump cuts and irrelevant content. For this reason, evaluation on AIR is particularly challenging as the models we test are image-based and there is no ground truth localising objects temporally. 
To test a video in AIR we sample two frames per second and rank the predictions obtained for each frame, aggregating the scores by taking the average of the top 5\% scores. 

\subsection{Results}

\begin{table}[t]
\centering
\resizebox{\columnwidth}{!}{%
\begin{tabular}{@{}l|ccc|ccc|c@{}}
\toprule
       & \multicolumn{3}{c|}{COFICUT}                      & \multicolumn{3}{c|}{VOST-AUG}                     & AIR~\cite{Moltisanti_2023_CVPR}            \\ \midrule
Model  & All            & Seen           & Unseen         & All            & Seen           & Unseen         & All            \\ \midrule
Random & 0.648          & 0.605          & 0.759          & 0.500            & 0.500            & 0.500            & 0.613          \\
CLS~\cite{Moltisanti_2023_CVPR}    & 0.692          & 0.684          & 0.723          & 0.625 & 0.631 & 0.589 & 0.623          \\
CLS\textsubscript{s}~\cite{Moltisanti_2023_CVPR} & 0.660 & 0.626 & 0.742 & \textbf{0.659} & \textbf{0.669} & \textbf{0.604}  & 0.619 \\
REG~\cite{Moltisanti_2023_CVPR}    & 0.722 & 0.702 & 0.778 & 0.575 & 0.584 & 0.538 & 0.621 \\
CANet~\cite{wang2023learning} & 0.710 & 0.686 & 0.811 & 0.492 & 0.487 & 0.535 & 0.617 \\
Ours   & \textbf{0.777} & \textbf{0.741} & \textbf{0.856} & 0.561          & 0.564          & 0.556          & \textbf{0.632} \\ \bottomrule
\end{tabular}%
}
\caption{Results obtained training models on VOST-AUG. The reported metric is MAP (Mean Average Precision) with macro averaging, where the two classes have equal weight. We report the performance of a random baseline which is equal to the support size of the positive class (the ``fine'' class). ``All/Seen/Unseen'' refers to performance evaluated respectively on all images and images showing objects seen/unseen in training.}
\label{tab:results}
\end{table}

\begin{table}[t]
\centering
\resizebox{\columnwidth}{!}{%
\begin{tabular}{@{}l|c|ccc@{}}
\toprule
Model                & Training dataset & All            & Seen           & Unseen         \\ \midrule
Random               & -                & 0.648 ± 0.000          & 0.605 ± 0.000          & 0.759 ± 0.000          \\
BCE                  & COFICUT          & 0.447 ±  0.026       & 0.447 ±  0.026       & -              \\
CLS~\cite{Moltisanti_2023_CVPR}                  & VOST-AUG         & 0.693 ±  0.040       & 0.688 ±  0.045       & 0.727 ±  0.091       \\
CLS\textsubscript{s}~\cite{Moltisanti_2023_CVPR} & VOST-AUG         & 0.665 ±  0.031        & 0.633 ±  0.030         & 0.751 ± 0.083        \\
REG~\cite{Moltisanti_2023_CVPR}                  & VOST-AUG         & 0.725 ±  0.017         & 0.706 ±   0.018       & 0.782 ± 0.053        \\
CANet~\cite{wang2023learning} & VOST-AUG & 0.713  ± 0.026 & 0.690 ± 0.030 & 0.811 ± 0.061 \\
Ours                 & VOST-AUG         & \textbf{0.779} ± 0.005 & \textbf{0.744} ± 0.025 & \textbf{0.856} ±  0.039\\ \bottomrule
\end{tabular}}
\caption{Results obtained with 5-fold cross validation on COFICUT. The reported metric is mean ± std MAP (classes have equal weight). Models trained on VOST-AUG were only tested on the five different folds, while BCE is a classification baseline where a model with the same backbone as the others is trained using the labels available on COFICUT.} 
\label{tab:results_coficut_train}
\end{table}

Table~\ref{tab:results} compares the performance of the models trained on VOST-AUG and tested on COFICUT, VOST-AUG, and AIR. We note that all models surpass the random baseline on all datasets (except CLS and CANet~\cite{wang2023learning} on some metrics and datasets), which validates our augmentation method: models can tell a coarsely cut object from a finely cut one after being trained on synthetic images without coarse/fine labels. Recall from Table~\ref{tab:coficut} that we generated VOST-AUG's training set from only 96 original images.
\vspace{5pt}

We highlight that COFICUT is the most appropriate benchmark for this task as it collects manually reviewed real images of cut objects with a significant visual domain gap. Despite such gap, the diversity of our augmented images allows us to successfully train a model to recognise coarseness in out-of-domain images. In particular, our model achieves the best results by a large margin. This is thanks to the auxiliary task introduced with the UNet decoder, which helps the backbone to focus on the minute details that distinguish the coarseness of a cut.
This is evident comparing the regression model (REG) with our model, since REG is essentially an ablation of our model where we discard the Decoder and the segmentation task.  
Our model is better than REG on the realistic datasets, COFICUT and AIR. 
This validates the idea of adding the extra task to provide auxiliary guidance. The performance gain for unseen objects further highlights the ability of our model (which is object-agnostic) to generalise well despite the visual gap. 

On VOST-AUG we note that CLS (training images split according to $c$) and CLS\textsubscript{s} (split according to seeding points) achieve the best performance. This is not surprising as the model is trained to separate images in the same (CLS) or similar (CLS\textsubscript{s}) way as they are split for testing. 
However, on the remaining datasets both CLS variants rank lowest. Performance on COFICUT unseen objects is even lower than the random baseline, which indicates that the model struggles to generalise.
This suggests that splitting images into two classes in our setting is a sub-optimal choice since
we have a continuum of simulated cuts ranging from very thin to very coarse, without a neat separation into two classes. 
We also note that CANet~\cite{wang2023learning} achieves decent results on COFICUT, but performs worse than the random baseline on VOST-AUG on the all/seen metrics. As mentioned before, attribute learning is a different task, so it is difficult for the model to perform well in our distinct setting. 

On AIR we observe that performance is poor across all models and closer to the random baseline. This is due to the fact that AIR is a video dataset, so without a ground truth localising the object it is difficult for any model to effectively predict the coarseness of the object shown in the video.

\paragraph{Training on COFICUT} We now check whether it would be possible to successfully train a binary classifier on COFICUT. We train the same backbone employed for the other models with binary cross entropy, using the labels available on COFICUT. Given its small size, we conduct this experiment with 5-fold cross validation, comparing against models trained on VOST-AUG by testing them on each fold as well. From Table~\ref{tab:results_coficut_train} we see that the model trained on COFICUT (``BCE'' in the Table) severely under-performs, with results well below the random baseline (there is no unseen MAP for BCE since all objects are now seen in training). This was expected as COFICUT contains in total 1,869 images, so the model overfits to the training set. However, this also shows that coarseness classification is not a trivial problem and that large training datasets are necessary. Instead of manually annotating images, our augmentation method allows to automatically generate a high-quality, large training dataset that models can successfully learn from. We also note that results with our methods are more robust as they exhibit a lower variation (MAP std is lowest).

\section{Conclusion}

\label{sec:conclusion}

We addressed the problem of recognising the end state of an action expressed by the manner in which it is performed. We explore this focusing on the cutting action, proposing an approach to detect whether an object is cut \textit{coarsely} or \textit{finely}.
We devise an effective image augmentation method to simulate an object being cut at different coarseness levels and in different ways. Starting from only 96 images, we were able to synthesise 47,395 images to train models to successfully recognise whether an object is cut finely or coarsely, without labels. Despite being trained on synthetic images, models achieve good performance on real images and even on unseen objects.
We also proposed a model to better leverage the data, boosting performance 
by over 4\%.

\paragraph{Limitations} Our augmentation method does not analyse the input scene, and as a result the synthesised image might sometimes look unrealistic. 
Also, objects may be cut while being held by hand mid-air, 
in which case the augmentation method produces an image of ``levitating'' object pieces, as we show in Section~\ref{sec:a-vost-more-details}. 
Scene understanding or affordance approaches~\cite{hassanin2021visual} could be employed to alleviate this issue. We also need a good object mask to synthesise good images. Recent segmentation models (e.g. Segment Anything~\cite{kirillov2023segment}) could help lifting this requirement, though objects would still need to be localised (e.g. providing a 2D point or a text prompt describing the object). 

\paragraph{Future Directions} 
Being object agnostic, our augmentation method can be adapted to synthesise images where the end state of an object affects its geometry and shape. 
For example, our method could be extended to predict the \textit{completeness} of a cut, i.e., telling whether an object is \textit{fully} or \textit{partially} cut. 
Other directions include adapting the augmentation method to synthesise videos 
and use the augmented data to instil knowledge in retrieval and generative models. 

\paragraph{Acknowledgements} Research funded by UKRI through the Edinburgh Laboratory for Integrated Artificial Intelligence (ELIAI) and the Turing Advanced Autonomy project.

\appendix

\section{Using Seeding Points as Regression Target}
\label{sec:a-seed-points-ablation}

\begin{table}[t]
\centering
\resizebox{\columnwidth}{!}{%
\begin{tabular}{@{}l|c|ccc|ccc|c@{}}
\toprule
                      & \multicolumn{1}{l|}{}                   & \multicolumn{3}{c|}{COFICUT} & \multicolumn{3}{c|}{VOST-AUG} & AIR~\cite{Moltisanti_2023_CVPR}   \\ \midrule
Model                 & L1 target                      & All     & Seen    & Unseen  & All     & Seen    & Unseen   & All   \\ \midrule
Random                & -                                      & 0.648   & 0.605   & 0.759   & 0.500   & 0.500   & 0.500    & 0.613 \\ \midrule
\multirow{2}{*}{REG~\cite{Moltisanti_2023_CVPR}} &
  Seed. p. & 0.670 & 0.640 & 0.762 & 0.501 & 0.501 & 0.503 & 0.603 \\
                      & \multicolumn{1}{l|}{Change r.} & 0.722 & 0.702 & 0.778 & 0.575 & 0.584 & 0.538 & 0.621 \\ \midrule
\multirow{2}{*}{Ours} & Seed. p.                         & 0.732   & 0.686   & 0.819   & \textbf{0.585}   & \textbf{0.602} & \textbf{0.571}    & 0.618 \\
 &
  \multicolumn{1}{l|}{Change r.} &
  \textbf{0.777} &
  \textbf{0.741} &
  \textbf{0.856} &
  0.561 &
  0.564 &
  0.556 &
  \textbf{0.632} \\ \bottomrule
\end{tabular}%
}
\caption{Results obtained training models on VOST-AUG with different regression targets: number of seeding points (Seed. p.) and the change ratio (Change r., $c$ in Equation 1 in the paper) we proposed to measure the coarseness of the simulated cuts.}
\label{tab:results_supp}
\end{table}

In this Section we validate the introduction of the change ratio value ($c$, see Equation 1 in the paper) used to define coarseness and as a regression target. 
As discussed in the paper, the number of seeding points controls the coarseness of the simulated cut, however other parameters involved in our augmentation method also affect the perceived coarseness. 
To show that using only the number of seeding points to measure coarseness is a sub-optimal choice, we train both our model and REG~\cite{Moltisanti_2023_CVPR} using the number of seeding points as target for the L1 loss instead of $c$. Table~\ref{tab:results_supp} compares results obtained with the two regression targets. Results obtained with seeding points as regression target are worse for both models on COFICUT and AIR, but better on VOST-AUG for our model. 
These results suggest that the alternative regression target limits the ability of the model to generalise to real images while overfitting to the training domain. We conclude that using the change ratio to gauge coarseness is thus a better way to train the model.

\section{VOST-AUG}
\label{sec:a-vost-more-details}

\begin{figure*}[t]
    \centering
    \includegraphics[width=0.8\linewidth]{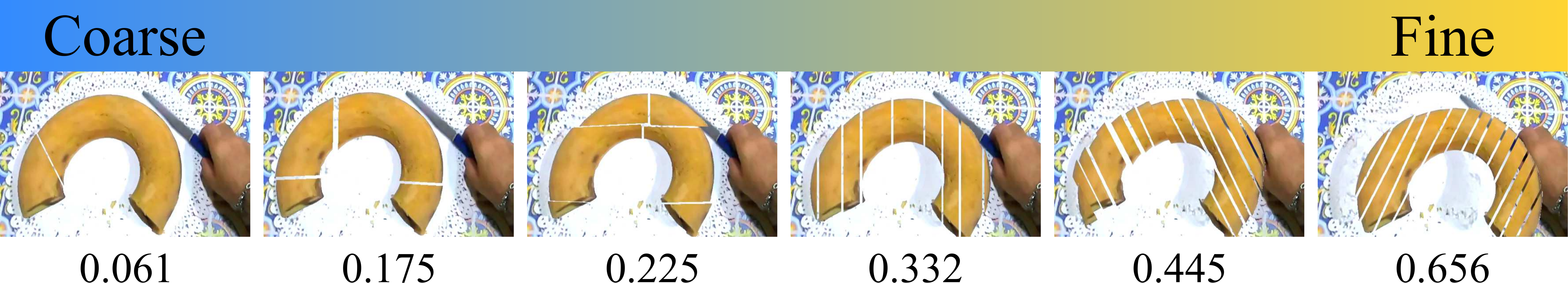}
    \caption{Showing how $c$ (reported at the bottom, see Equation 1 in the paper) varies for a set of images augmented from the same source.}
    \label{fig:c_values}
\end{figure*}
\begin{figure*}
    \centering
    \includegraphics[width=0.8\linewidth]{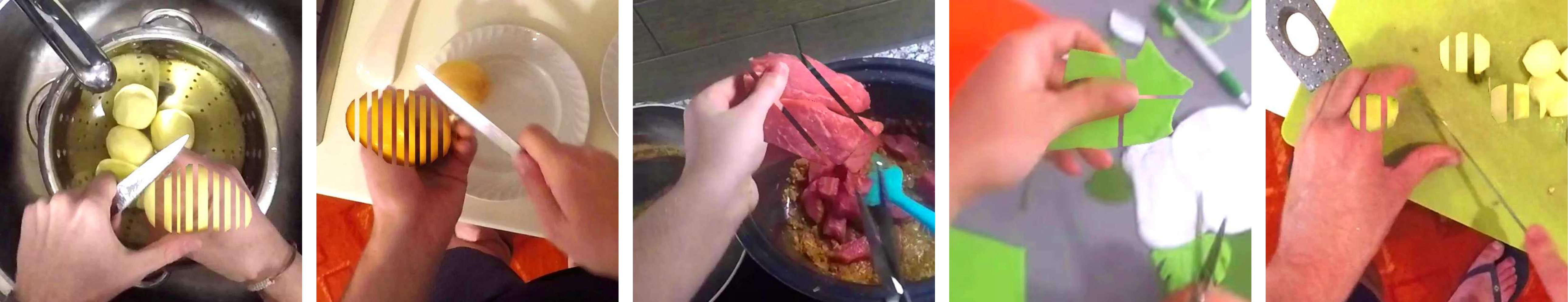}
    \caption{Examples of failure cases of our augmentation method. When objects are cut while held mid-air the simulated cuts look unrealistic. In some cases object parts are pushed onto hands or other objects.}
    \label{fig:fail_cases}
\end{figure*}

\paragraph{Illustrating $c$ Values}Figure~\ref{fig:c_values} illustrates how $c$ (see Equation 1 in the paper) varies for a set of images synthesised from an original image. Note how small/large values visually correspond to a coarser/finer cut.

\begin{figure*}[t]
    \centering
    \includegraphics[width=\linewidth]{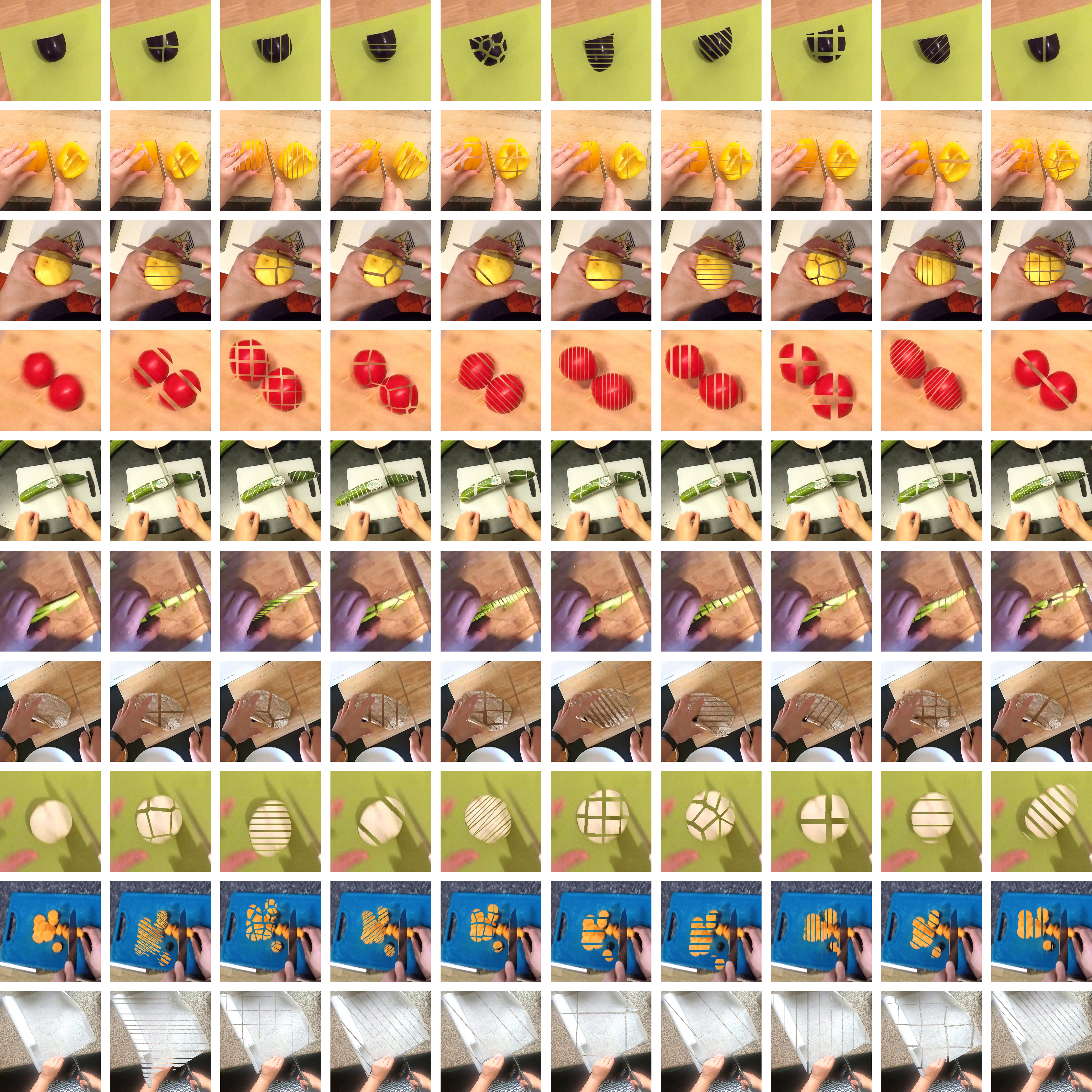}
    \caption{Examples from VOST-AUG. We show a few images augmented from a single source (left-most column). Images are cropped to improve visualisation. Best seen zoomed-in on a screen.}
    \label{fig:vost_examples}
\end{figure*}

\paragraph{Failure Cases} Figure~\ref{fig:fail_cases} illustrates examples where our augmentation method fails to synthesise realistic images. This happens mostly when objects are cut while held mid-air, which causes the split object to appear as though it ``levitates''. In some cases objects regions are pushed over hands or other objects, which also simulates a less realistic image. As noted in the paper, these issues could be alleviated using scene understanding or affordance models.

\paragraph{More Examples} Figures~\ref{fig:vost_examples} shows more synthesised images from VOST-AUG, together with the corresponding original source (left-most column). To facilitate illustration we crop images around the object. Our method works well in challenging conditions, e.g., when the original image shows more than one instance of the object or when the object is held in hand (third row from the top). The augmentation method is able to generate good images regardless of the object size and shape
We note that the majority of images simulates realistic cuts, though as the number of seeding points increases (i.e., as the number of split parts increases), images may tend to look more artificial. This is not a concern as the purpose of these images is to train a model, which we are able to do successfully as demonstrated in the paper.

\paragraph{Seen and Unseen Objects} The objects \textbf{seen} during training in the VOST-AUG train split are: ``aubergine, beef, bread, broccoli, butter, cake, carrot, chicken, chilli, cloth, courgette, cucumber, dough, garlic, ginger, gourd, guava, lettuce, mango, olive, onion, paper, pea, peach, pepper, potato, pumpkin, salad, tomato, vegetable''. The \textbf{unseen} objects are: ``asparagus, bacon, celery, corn, ham, herbs, ladyfinger, melon, mozzarella, spinach, spring onion''. 

\section{COFICUT}
\label{sec:a-coficut-details}

\begin{figure*}[t]
    \centering
    \includegraphics[width=\linewidth]{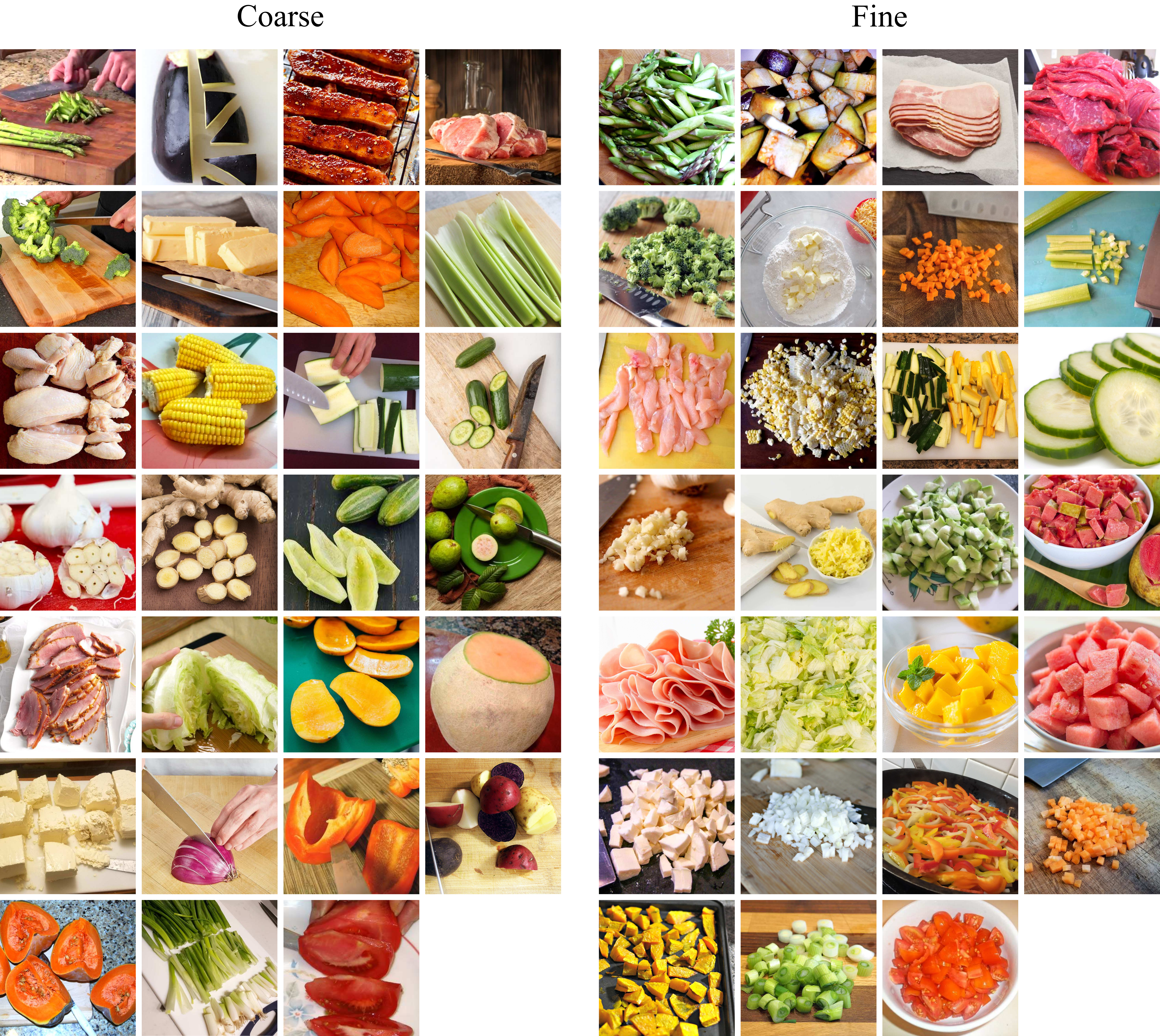}
    \caption{Examples from the COFICUT evaluation dataset. We show one coarse/fine image for each object. From top-left: ``asparagus, aubergine, bacon, beef, broccoli, butter, carrot, celery, chicken, corn, courgette, cucumber, garlic, ginger, gourd, guava, ham, lettuce, mango, melon, mozzarella, onion, pepper, potato, pumpkin, spring onion, tomato''.}
    \label{fig:coficut_more_examples}
\end{figure*}

The list of objects in COFICUT after reviewing is: ``asparagus, aubergine, bacon, beef, broccoli, butter, carrot, celery, chicken, corn, courgette, cucumber, garlic, ginger, gourd, guava, ham, lettuce, mango, melon, mozzarella, onion, pepper, potato, pumpkin, spring onion, tomato''. 
Amongst these, the following were not seen during training: ``asparagus, bacon, celery, corn, ham, melon, mozzarella, spring onion''.
Figure~\ref{fig:coficut_more_examples} shows more images from COFICUT (one coarse/fine per object). Note the diversity of the images (point of view, lighting, style), especially compared to the training images from VOST-AUG, and how distinct each object looks in its coarse and fine states.

\section{More Examples from InstructPix2Pix}
\label{sec:a-more-examples-p2p}

\begin{figure*}[t]
    \centering
    \includegraphics[width=\linewidth]{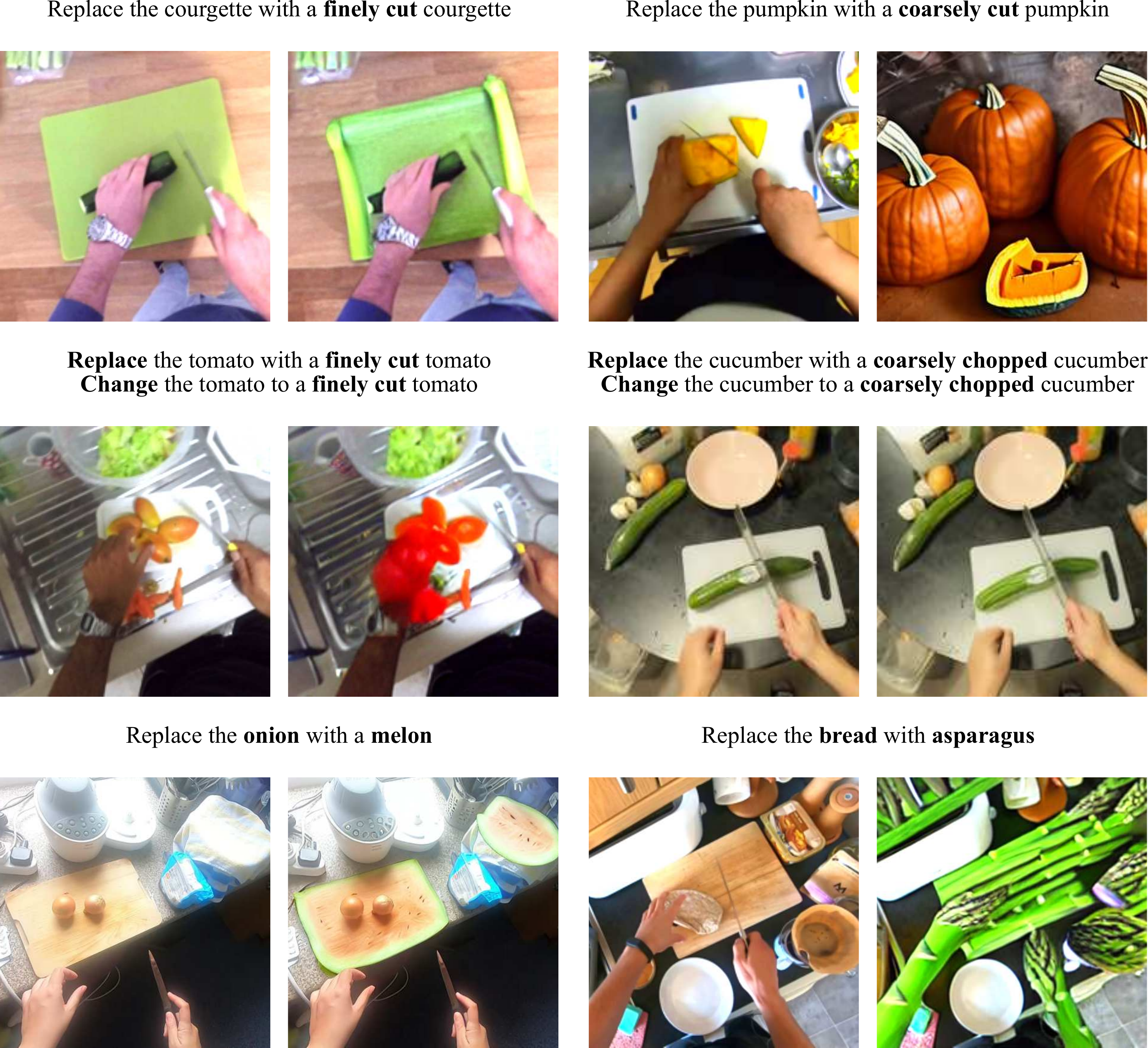}
    \caption{Examples from our experiments with InstructPix2Pix~\cite{brooks2023instructpix2pix}. Text indicates the prompts used.}
    \label{fig:more_probing}
\end{figure*}

Figure~\ref{fig:more_probing} shows more examples from our experiments with InstructPix2Pix~\cite{brooks2023instructpix2pix}. As seen in the paper, the model ignores the adverb specified in the prompt and fails to replace the indicated object with another one in a realistic way, often hallucinating the image. We speculate that the model relies heavily on colour to ground the queried object to the image. 
We thus hypothesise that the model struggles to separate the object when it has a similar colour to its surrounding elements. This is particularly visible in the bottom right example in Figure~\ref{fig:more_probing}, where the bread and the whole scene share a similar colour. Note how the model inpaints asparagus over the whole image, including the hands and arms of the subject, the chopping board and the cupboard.

In many cases the model did not modify the input image at all. We do not illustrate these cases here. We show in Figure~\ref{fig:more_probing} (middle row) that results are independent of the prompts wording, i.e., we obtained the same results when changing the words of the prompt. 

{
    \small
    \bibliographystyle{ieeenat_fullname}
    \bibliography{main}
}

\end{document}